# FishMOT: A Simple and Effective Method for Fish Tracking Based on IoU Matching

Shuo Liu[a,b,c], Lulu Han[a,b,c], Xiaoyang Liu[a,b,c], Junli Ren[a,b,c], Fang Wang[a,b,c], Ying Liu[c,d], Yuanshan Lin[a,b,c]

a. School of Information Engineering, Dalian Ocean University, Dalian 116023, China
b. Liaoning Key Laboratory for Marine Information Technology, Dalian 116023, China
c. Key Laboratory of Environment Controlled Aquaculture, Ministry of Education, Dalian Ocean University, Dalian 116023, China
d. College of Biosystems Engeering and Food Science, Zhejiang University, Hangzhou 310030, China

This study was partly supported by Foundation of the Liaoning Educational Committee (LJKZ0730, QL202016), Applied Basic Research Programs of Science and Technology Commission Foundation of Liaoning Province(2022JH2/101300187), Open Foundation of the Key Laboratory of Environment -Controlled Aquaculture (Dalian Ocean University) Ministry of Education (202219), the Central Government Subsidy Project for Liaoning Fisheries (2023) and Liaoning Province Natural Science Foundation (2020-KF-12-09).

**Abstract**: Fish tracking plays a vital role in understanding fish behavior and ecology. However, existing tracking methods face challenges in accuracy and robustness dues to morphological change of fish, occlusion and complex environment. This paper proposes FishMOT(Multiple Object Tracking for Fish), a novel fish tracking approach combining object detection and IoU matching, including basic module, interaction module and refind module. Wherein, a basic module performs target association based on IoU of detection boxes between successive frames to deal with morphological change of fish; an interaction module combines IoU of detection boxes and IoU of fish entity to handle occlusions; a refind module use spatio-temporal information uses spatio-temporal information to overcome the tracking failure resulting from the missed detection by the detector under complex environment. FishMOT reduces the computational complexity and memory consumption since it does not require complex feature extraction or identity assignment per fish, and does not need Kalman filter to predict the detection boxes of successive frame. Experimental results demonstrate FishMOT outperforms state-of-the-art multi-object trackers and specialized fish tracking tools in terms of MOTA, accuracy, computation time, memory consumption, etc.. Furthermore, the method exhibits excellent robustness and generalizability for varying environments and fish numbers. The simplified workflow and strong performance make FishMOT as a highly effective fish tracking approach. The source codes and pre-trained models are available at: https://github.com/gakkistar/FishMOT

# 1. Introduction

One of the challenges of aquaculture is to monitor and manage the behavior and growth of fish. Fish tracking technology is a key solution for this challenge, as it enables the collection and analysis of data on fish movement, activity, and size. Fish tracking technology can help improve the performance and sustainability of aquaculture systems, as well as the quality and safety of the fish products(Zhou et al., 2018). Fish tracking and behavior analysis can also be used to keep an eye on the environment that supports fish growth, such as water quality, temperature, and oxygen levels(Xia et al., 2018).This can provide valuable insights for improving fish welfare, productivity, and profitability(Yang et al., 2021). Fish tracking technology can be implemented using various sensors and methods. As a non-invasive, objective and repeatable tool, machine vision has been a widely used tool for fish tracking in aquaculture(An et al., 2021).

Fish tracking using machine vision can provide useful information for fisheries management, aquaculture, and marine ecology. However,fish are non-rigid objects that can change their shape and posture due to swimming, bending, or twisting. This can affect their characteristics and appearance model, and easily cause false detection or tracking by the tracker(Zhou et al., 2017). Furthermore, when fish move relative to each other, they can occlude each other partially or totally. This can lead to wrong fish identity association and tracking failure. Last but not least, the quality and visibility of the images or videos can be affected by the complex environment that fish live in, such as uneven illumination, water turbidity, reflection, refraction, and scattering. This may create false positives or negatives in the fish detection and then lead to tracking failure. In a word, fish tracking using machine vision still faces many difficulties, such as morphological change of fish, occlusion, and complex environment.

To this end, this paper proposes a novel tracking approach termed FishMOT (Multiple Object Tracking for fish) that leverages the speed and accuracy of multi-fish tracking. This approach matches fish identities only employing IoU (Intersection over Union) of detection boxes in successive frames and IoU of fish entities, which overcome the difficulties resulting from morphological change of fish and occlusion between each other. Additionally, it uses spatio-temporal information of fish to handle the tracking failure resulting from the missed detection by the detector under complex environment. This method significantly reduce the requirements of computation time and storage space. Experimental results demonstrate that FishMOT achieves much higher accuracy and tracking speed than the state-of-the-art multi-object tracking methods and existing fish tracking methods. And FishMOT also performs well for tracking in different environments.

The main contributions of this paper are as follows:
- We propose a new fish detection-based tracking method **FishMOT**, which achieves higher tracking accuracy and speed than those of famous fish tracking methods idtracker.ai (Romero-Ferrero et al.,2019) and TRex(Walter et al., 2021) on the idtracker.ai dataset. Besides, FishMOT outperforms existing state-of-art multi-object tracking methods.

- We develop a strategy that combines IoU of detection boxes in successive frames and IoU of fish entities to handle the occlusion and morphological change of fish.
- We design a strategy that uses the spatio-temporal information of fish individuals to overcome the tracking failure resulting from the missed detection by the detector under complex environment.

## 2. Related work

### 2.1 Object detection

Object detection is a computer vision task that involves locating and identifying one or more objects in an image. Object detection has many applications, such as face recognition, self-driving cars, surveillance, and image retrieval. Since deep learning can use neural networks to learn from large amounts of data and perform complex tasks, deep learning has been widely used for object detection in recent years, due to its high accuracy and speed(Xiao et al., 2020). YOLO (You Only Look Once) is a family of real-time object detection algorithms (Redmon et al., 2016) that use a deep convolutional neural network (CNN) to predict bounding boxes and class probabilities for each object in an image. YOLOv7 (Wang et al., 2022) is a new version of the YOLO object detection algorithm. It supports multi-scale training and testing, which means that it can handle images of different sizes and resolutions without losing accuracy or speed. And it also supports dynamic input resolution, which allows it to adjust the input size according to the available GPU memory. Moreover, it requires several times cheaper hardware than other neural networks and can be trained much faster on small datasets without any pre-trained weights. In a word, YOLOv7 is expected to become the industry standard for object detection in the near future. In this paper, we use YOLOv7 as the detector of FishMOT.

### 2.2 Multiple object tracking

Multiple object tracking (MOT) is a challenging computer vision task that aims to locate and identify multiple objects of interest in a video sequence and maintain their temporal consistency(Luo et al., 2021) MOT has many applications in various domains, such as surveillance(Ahmed et al., 2021), autonomous driving(Ravindran et al., 2020), sports analysis(Romeas et al., 2016), and human-computer interaction(Yi et al., 2022). However, directly applying existing MOT to track fish with high appearance similarities and irregular motion (Chen et al., 2021) fares poorly. One of the most popular and effective frameworks for MOT is tracking by detection(Zhang et al., 2022), which consists of two main steps: object detection and data association. Object detection is the process of finding and classifying the objects of interest in each frame of the video. Data association is the process of linking the detected objects across frames to form consistent trajectories. Tracking by detection has several advantages over other frameworks, such as being able to handle new objects entering or leaving the scene, being robust to noise and outliers, and being able to leverage the advances in deep learning for object detection(Wang et al., 2022). To this end, FishMOT uses the framework of tracking by detection. However, tracking by detection also has some limitations and challenges, such as requiring high-quality object detectors, being sensitive to detection errors and missed detections, and having high computational complexity and memory consumption(Rakai et al., 2022). To overcome these limitations, we developed several strategies in the data association according to the characteristics of fish and the characteristics of surrounding environment that fish live in. Among them, the basic module and the interaction module mainly address the shortcomings of high computational complexity and large memory consumption; the refind module mainly solves the problem of tracking algorithm being sensitive to detection errors and missed detections of the detector.

**2.3  Fish tracking**

Fish tracking is a valuable technique for various purposes, such as studying fish behavior, ecology, monitoring fish health and welfare, managing fisheries and aquaculture, and conserving endangered species. Different methods can be used for fish tracking, such as traditional probabilistic texture analysis, deep learning methods based on neural networks, and multiple object tracking methods based on data association.

Traditional probabilistic texture analysis methods for fish tracking extract texture features from fish images and compare them with a probabilistic model. Some examples of these methods are Idtracker(Pérez-Escudero et al.,2014), ToxTrac(Rodriguez et al.,2018), and automated planar tracking(Rasch et al., 2016). Idtracker uses a complex algorithm based on a Bayesian analysis with a similarity metric to compare the objects' texture. ToxTrac, on the other hand, uses a combination of a similarity analysis with a Hungarian algorithm to manage the identity preservation of multiple fish. Similarly, automated planar tracking can also handle simultaneous tracking in multiple arenas, resulting in one of the most flexible free tracking tools for trajectory generation. However, these traditional probabilistic texture analysis methods have limitations in tracking many targets or for long durations.

Deep learning (DL) methods for fish tracking use neural networks to learn features and representations of fish from large amounts of data. DL-based fish tracking methods can achieve high accuracy and robustness in fish tracking and usually outperform those based on traditional probabilistic texture analysis. Fish CnnTracker(Panadeiro et al., 2021) is a DL-based fish tracking, which uses an approach based on deep learning models called "convolutional neural networks" (CNNs). It can track the position and orientation of multiple fish in a video by using CNNs to detect the fish regions and estimate their angles. Idtracker.ai and Trex are currently state-of-the-art, yielding high-accuracy in maintaining consistent identity assignments across entire videos. They are achieved by training an artificial neural network to visually differentiate between individuals, and using identity predictions from this network to avoid/correct tracking mistakes. Both approaches work without human supervision, and are up to approximately 100 unmarked individuals. They are currently available tools with visual identification for such large groups of individuals, and with such high quality of results. However, these methods usually demand significant computing resources and memory space for per-individual feature extraction. And their accuracy are affected by video quality, lighting conditions, individual appearance differences and other factors.

On the other hand, multiple object tracking methods can handle complex scenarios. So, some researchers attempt to perform fish tracking by using multiple object tracking methods. Wang et al. (2022) proposed YOLOv5 and SiamRPN++ networks for detecting and tracking abnormal fish behaviors in aquaculture. However, they do not account for environmental complexities like lighting, water currents, or pollution that may compromise stability and robustness. Li et al. (2022) presented CMFTNet based on FairMOT (Zhang et al., 2021) and CenterNet (Zhou et al., 2020), which can detect and track the movement trajectory of fish schools in complex water environments. In the backbone part of the network, deformable convolutional network is used to enhance the contextual features of fish to adapt to the non-rigid deformation and scale changes of fish. However, this method uses a model based on joint detection embedding paradigm, which cannot achieve the high accuracy of detection and association methods. The main difficulties of multiple object tracking for fish are as follows: fish have diverse shapes, large color changes, fast movement, easy to occlude and cross, once the features are lost, the tracking of fish will fail; the underwater environment is complex, the illumination is uneven, the water quality is turbid, resulting in low image quality, high noise, low contrast (Li et al.,2014; Mei et al.,2022); the number of fish is uncertain, may appear or disappear at any time, need to dynamically adjust the number and position of tracking targets (Hernández et al.,2018); the behavior and group characteristics of fish need to be considered, such as the density, direction, distribution of fish schools (Wang et al.,2017).

## 3. Methods

This paper proposes a simple yet effective fish tracking approach that utilizes IoU of detection box and IoU of fish entity for data association. It handles adjacent or crossing fish without requiring additional algorithms like Kalman filters or feature extraction, thereby improving efficiency and accuracy. This section details the tracking methodology.

### 3.1 Overview

The overview of FishMOT is shown in **Fig. 1**. Since YOLOv7 can balance detection accuracy and speed, we use it as the detector. After obtaining the position data of the fish, the algorithm will first judge whether there is any missing detection by the detector. The basic module of FishMOT is the main module of the algorithm, this module will calculate the IoU value of all detection boxes. If all detection boxes have a high IoU value, it will directly perform the first ID association and then complete the allocation; if some detection boxes have two or more high IoU values, it will be processed by the interaction module, which will obtain the image in the detection box and binarize it, then obtain the largest shadow connected area of the image, calculate the IoU value of the fish entity, and combine the IoU of the bounding box and the entity IoU to complete the ID allocation through the second association. If there are any missing targets, it will enter the refind module to deal with them. This module will combine the supplementary temporary data and the spatio-temporal information of the missing targets to associate the target ID, and finally calculate the approximate position of the missing targets after completing the ID allocation. After completing the temporary data supplement by the refind module, the algorithm will calculate the IoU of the bounding box of the detection targets in the previous and next frames according to the corresponding relationship between the two frames. **Fig. 2** depicts the tracking trajectories, each fish uses a random id number and trajectory of the same color.

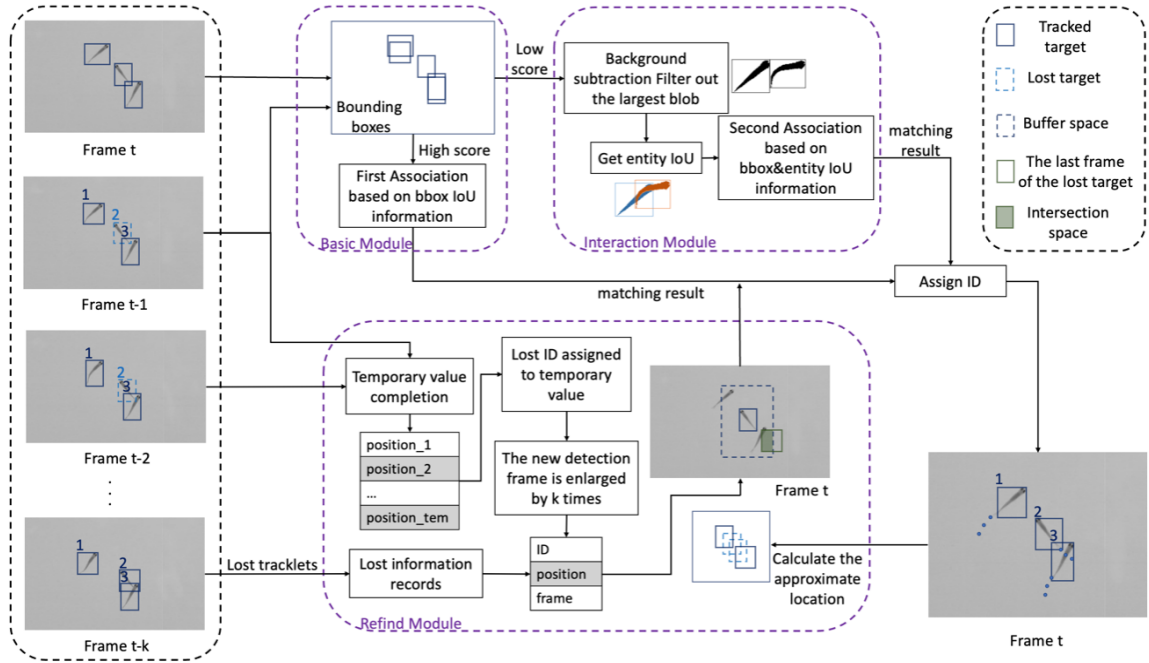

**Fig. 1.** Overview of FishMOT. Once the original video is input to FishMOT, detection boxes can be obtained by YOLOv7, then they go through Basic Module, Interaction Module and Refind Module, and finally completes ID assignment via IoU-based association.

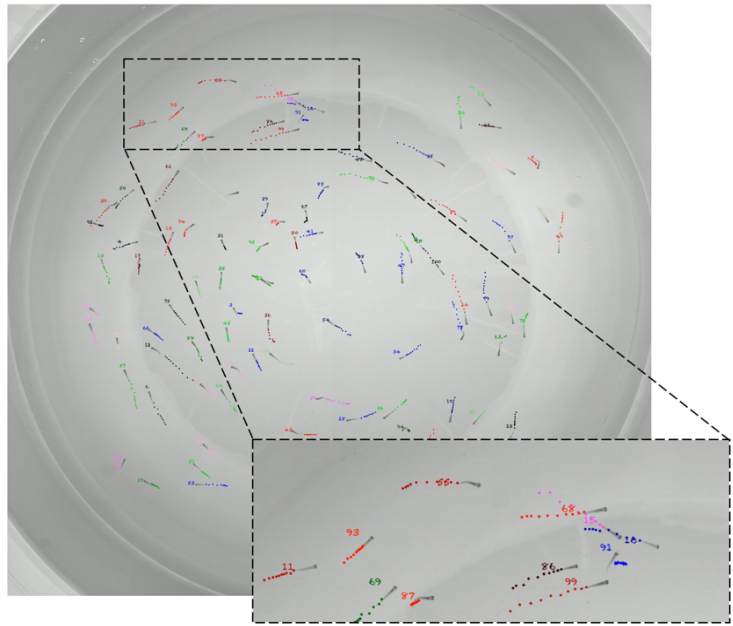

**Fig. 2.** The fish trajectories generated by FishMOT.

### 3.2 Basic Module

Target association is a crucial step for multi-object tracking based on detection, which has a direct impact on the tracking performance. The common approach is: given the current detection bounding box, apply the Kalman filter to predict the location of the bounding box in the next frame, and then associate the target based on the predicted value of the bounding box in the next frame and the intersection-over-union (IoU) value of the actual detection bounding box in the next frame.

Traditional fish movement tracking employs Kalman filters to conduct motion prediction and tracking. The Kalman filter is a widely used technique for state estimation of linear dynamic systems, which can update the optimal estimate and error covariance matrix of the state based on the observation data and system model. The benefit of the Kalman filter is that it can efficiently handle noise and uncertainty, and enhance the precision and stability of prediction.

However, several issues may arise when using Kalman filters for fish motion prediction: First, Kalman filters require prior knowledge of the system and noise models, but fish motion may be influenced by various factors such as water currents, food sources, companions, etc. These factors may be hard to measure or model, leading to inaccurate or incomplete system and noise models that affect the prediction quality. Second, Kalman filters assume linearity in the system, but fish motion may exhibit nonlinearity such as turning, accelerating or decelerating. These motions may invalidate the linearity assumption of Kalman filters and compromise the prediction quality. As **Fig. 3** illustrates, fish can abruptly alter their motion direction while Kalman filter continues to predict based on the previous motion direction. This results in a large offset between the predicted and actual positions. Lastly, Kalman filters demand high computational complexity and storage space especially when the state and observation dimensions are high. These resources may be constrained or insufficient and hamper the prediction efficiency.

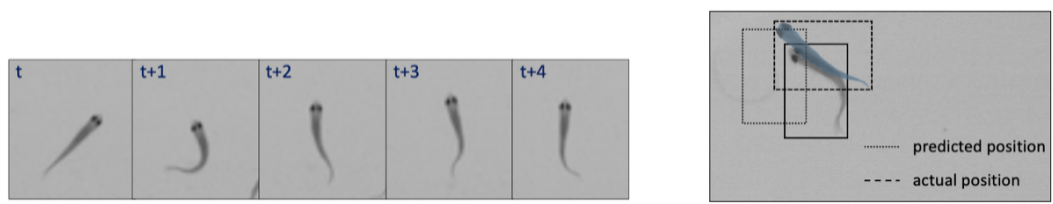

(a) The 5 Intra-frame fish pose changes    (b) The predicted and actual position

**Fig. 3.** The motion posture samples and fish prediction change samples. It is evident that there is a significant discrepancy between the predicted and actual positions. It can also be seen that the randomness of fish movement.

On the other hand, we observed fish groups carefully and found that although the shape and orientation of the individual fish in each frame are quite different, the position of each fish in adjacent frames do not differ much. To verify our guess, we analyze the videos from idtracker.ai in term of IoU of adjacent frames. The results are shown in **Fig. 4**. Where (a) and (b) are the average IoU of adjacent frames and the distribution of all IoU values over 10 fish; (c) and (d) are the average IoU value of adjacent frame and all IoU values over 100 fish. It can be seen that the IoU of adjacent frames maintains a high value, about 0.6, indicating little change of fish's position in adjacent frames. Therefore, we believe that the IoU of adjacent frames can be used for target association.

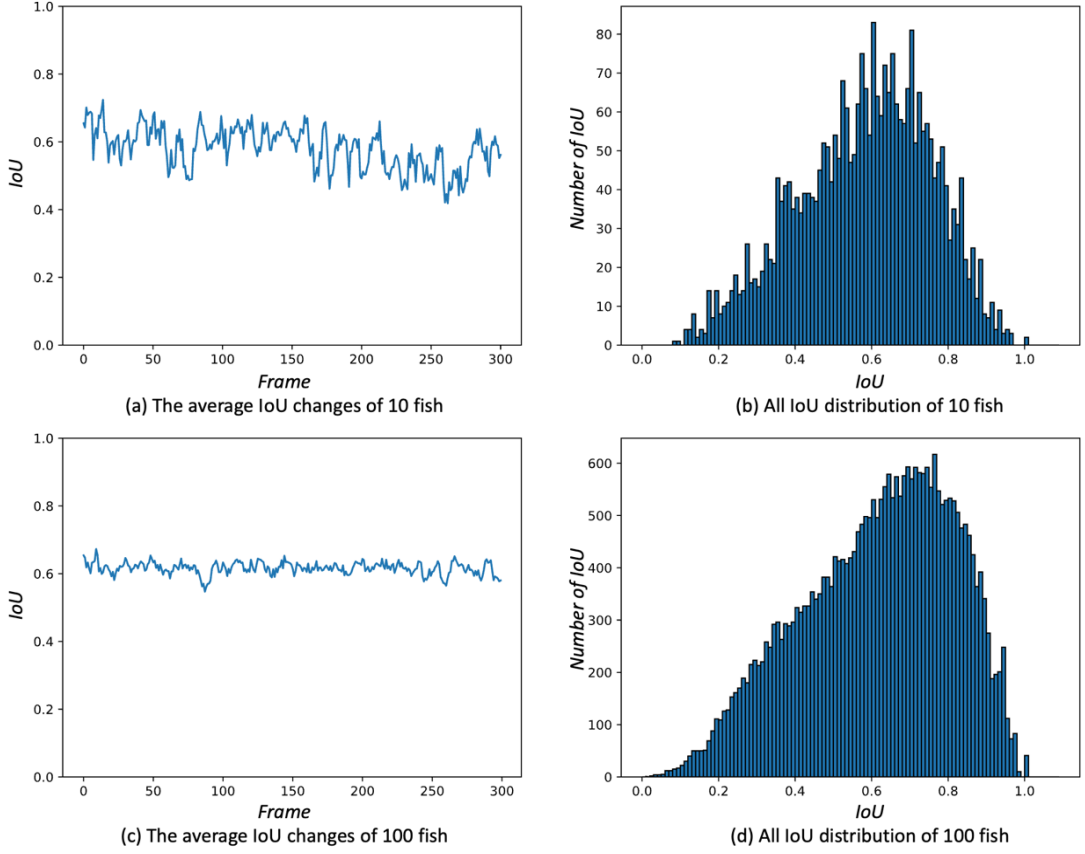

**Fig. 4.** IoU variation curve and IoU distribution histogram. The average IoU changes variation curve of 10 and 100 fish and all IoU distribution histogram of 10 and 100 fish.

In fact, the IoU of adjacent frames characterize the transient spatio-temporal information of individual motion. Particularly in scenarios where individuals have similar appearances, such short-term spatio-temporal information becomes exceedingly vital for target tracking. Hence, the IoU of adjacent frames should be as a potential strategy for target association. Its calculation formula is given in **Eq. (1)**

$$iou(d_t^i, t_{t-1}^j) = \frac{area(d_t^i \cap t_{t-1}^j)}{area(d_t^i \cup t_{t-1}^j)} \quad (1)$$

Based upon the foundation of adjacent frame IoU, we design a basic module to calculate the degree of overlap between detection boxes across consecutive frames, obtaining the IoU and utilizing the Hungarian algorithm for matching. If each detection box maintains a unique IoU value with the preceding frame, ID association and allocation can be conducted directly, thus facilitating fish tracking between continuous frames; in cases where multiple high IoU values are present, the algorithm will engage the interactive module for more precise IoU computation.

### 3.3 Interaction Module

Target association based on IoU of detection boxes between successive frames for target association works well for fish tracking in most cases. However, when more than two fish approach or cross each other, they may enter a state of occlusion in the top camera's view., as **Fig. 5**. In this situation, relying only on the IoU of detection boxes for matching may result in ID switching among the fish.

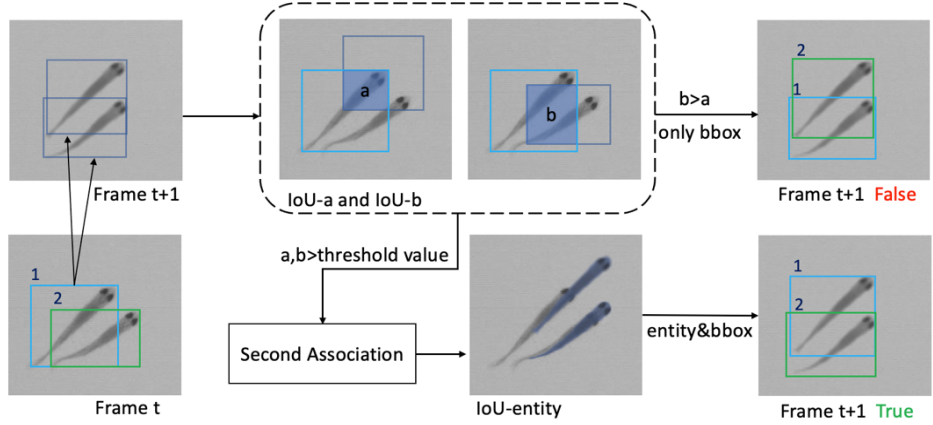

**Fig. 5.** The samples of fish approaching each other using bboxes and entities.

To handle the challenges of occlusion and morphological change of fish, we propose an interaction module that combines the IoU of detection boxes and fish entities for target association (see **Fig. 5**). The interaction module employs a two-step strategy to address this issue. First, it sets an IoU threshold for the potentially crossing fish. If two or more IoU values exceed the threshold, it does not use the IoU of detection boxes alone for ID matching. Second, it segments the fish entities within the detection boxes using background subtraction. This technique extracts the fish entities by binarizing the image within the bounding box and obtaining the largest connected region of the shadow part. Then, it computes the IoU of fish entities and combines it with the IoU of detection boxes for a second ID association. After the association is completed, it assigns IDs to the fish.

### 3.4 Refind Module

The inter-frame IoU based tracking depends heavily on the accuracy of detector. The detector may miss some targets when encountering some intractable situations, such as uneven illumination, water reflection, and so on. This may cause inconsistency in number of fish across frames and pose challenges for subsequent computation and matching processes. It may also lead to possible omissions during ID assignment resulting in unequal number of tracked fish in subsequent frames. To enhance fish tracking accuracy, handling omissions becomes essential.

We leverage inter-frame IoU correlation property and propose a refind module to cope with omissions. The buffer space of the module is shown in **Fig. 6.** This module effectively resolves target omission problem. Specifically, when a certain fish is successfully identified in current frame and it is not at identification area boundary, but it is not detected in subsequent frames that contain this fish, the algorithm temporarily uses provisional data for supplementation.

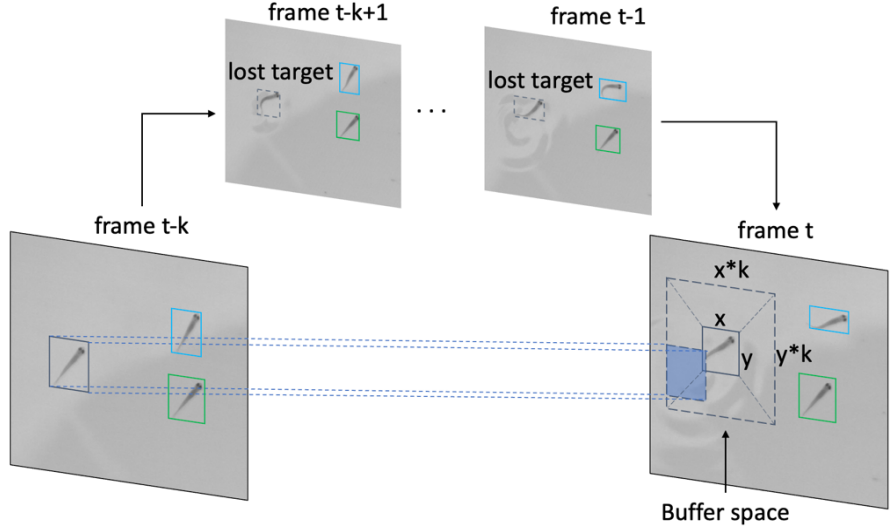

**Fig. 6.** The diagram of buffer space for handling the omissions of detector

After matching completion, IDs that have not been identified are assigned to provisional data. Then, within k frames, if a suitable candidate fish emerges to match the omitted fish, the algorithm assigns this omitted ID to this fish. Candidate fish matching must satisfy two conditions: First, they cannot be taken by other IDs, and second they must be within a buffer region with radius of k times detection box length and width centered on reserved ID location.

To prevent erroneous matching we also stipulate that reserved ID set must be a subset of unassigned ID set in subsequent frames. If this ID has not been successfully assigned after k frames, the algorithm discards this reserved ID. After successful matching, the algorithm estimates the box position and size of detections for omitted frames by using linear interpolation method. Then, the algorithm adds these estimated detection boxes to detection information to complete recovery of lost trajectory for omitted fish.

## 4. Experiments

We conduct four experiments to evaluate the performance of FishMOT. In Section 4.1, we describe the datasets, evaluation metrics, experimental settings, and parameter selections. In Section 4.2, we compare FishMOT with other state-of-the-art multi-object tracking algorithms on the validation sets of 10 and 100 fish, using the same detection model. In Section 4.3, we benchmark FishMOT against idtracker.ai and TRex on varying fish counts, demonstrating excellent performance.

### 4.1 Datasets and Evaluation Metrics

We utilized open-source video data from idtracker.ai, a dataset harvested from an overhead perspective encompassing fish video data. The delineation of bounding boxes and the training of the detection model were conducted using videos featuring 10, 60, and 100 fish. The tracking performance was evaluated using additional videos of 10 and 100 fish, for which supplementary ID annotations were carried out. Throughout the process, approximately 4,000 fish images were annotated within the training and testing sets, encompassing roughly 210,000 individual fish entities.

We use the following evaluation MOT metrics(Bernardin et al., 2008) to measure the performance of FishMOT:

Multiple Object Tracking Accuracy (MOTA): This metric combines three error sources: false positives, false negatives, and ID switches. MOTA will be calculated with **Eq. (2)**. Where $FN_t$, $FP_t$, $IDSw_t$ are the False Negative, False Positive, the number of id switches during tracking respectively for time t. It ranges from -∞ to 100%, where higher values indicate better performance.

$$MOTA = 1 - \frac{\sum_t(FN_t+FP_t+IDSw_t)}{\sum_t GT_t} \quad (2)$$

IDF1 Score (IDF1): This metric is the harmonic mean of ID precision and ID recall. It ranges from 0 to 100%, where higher values indicate better identification. IDF1 Score is a metric that measures the accuracy of identity assignment in multiple object tracking (MOT). It is the ratio of correctly identified detections to the average of ground truth and computed detection numbers. It is computed as **Eq.(3)**. Where IDTP is the number of detections with correct identity, IDFP is the number of detections with wrong identity, and IDFN is the number of missed detections with correct identity. A higher IDF1 Score indicates a better tracking performance, as it means the tracker can preserve the identity of the objects over time.

$$IDF1 = \frac{2*IDTP}{2*IDTP+IDFP+IDFN} \quad (3)$$

The IDP (Identification Precision) illustrates the ratio of correctly identified targets over all recognition results; a higher value denotes superior recognition precision, as delineated in **Eq.(4)**. Meanwhile, the IDR (Identification Recall) expresses the proportion of accurately identified targets to the entire set of authentic targets; a higher figure signifies a greater recognition recall rate, as depicted in **Eq.(5)**. In multi-object tracking systems, it is imperative to maximize IDP to reduce false positives and to elevate IDR to diminish missed detections, thereby enhancing the overall system performance. Typically, a balance between IDP and IDR must be struck.

$$IDP = \frac{IDTP}{IDTP+IDFP} \quad (4)$$

$$IDR = \frac{IDTP}{IDTP+IDFN} \quad (5)$$

**4.2 Performance comparison of multi-object tracking algorithms**

We compared FishMOT with other state-of-the-art multi-object tracking algorithms, using the same detection model. These algorithms include Sort (Bewley et al.,2016), DeepSort (Wojke et al.,2017), StrongSort (Du et al.,2023), UAVMOT (Liu et al.,2022), DeepMOT (Xu et al.,2020), ByteTrack, BotSort (Aharon et al.,2022), C_BIoU (Yang et al.,2023) and CMFTNet. Sort uses Kalman filter prediction and Hungarian algorithm for IoU matching. DeepSort adds an appearance discrimination network on top of Sort. StrongSort proposes two lightweight, plug-and-play, model-agnostic, appearance-free algorithms to further optimize the tracking results. UAVMOT designs a gradient balance focal loss function to supervise the learning of target heatmaps, which can handle small-scale targets in videos. DeepMOT proposes an end-to-end multi-object tracking training framework based on deep Hungarian network and CLEAR-MOT metrics, which can directly optimize the performance of DeepMOT methods. ByteTrack removes the Re-ID model and adds confidence to solve the problem of low-confidence score boxes caused by occlusion. BotSort modifies the Kalman filter to achieve better tracking effect. C_BIoU proposes a buffer intersection over union method, which is a geometric consistency measure for tracking multiple objects with indistinguishable appearance and irregular motion. CMFTNet introduces deformable convolutional network in the backbone part of the network, which enhances the contextual features of fish in complex environments and adapts to the non-rigid deformation characteristics of fish.

We first compare FishMOT with classical MOT algorithms on the 10-fish video. **Table 1** shows their comparison results. It can be seen from the table that most of the MOT algorithms do not perform well. CMFTNet performs the worst, with a MOTA of only 31.4%, while the famous DeepSort has a MOTA of only 36.13%. Among the these classical MOT algorithms, C_BIoU has the best performance, with 0 ID switches and a MOTA of 93.4%. The proposed FishMOT has better performance than the existing MOT algorithms in all metrics, whose MOTA is more than 6% higher than those of C_BIoU. This shows that FishMOT achieves state-of-the-art performance in fish tracking.

**Table 1** The comparison with other state-of-the-art MOT algorithms on the 10-fish video.

| Methods | IDSw↓ | IDP(%)↓ | IDR(%)↓ | IDF1(%)↑ | MOTA(%)↑ |
|---|---|---|---|---|---|
| CMFTNet | 155 | 37.08 | 34.4 | 35.69 | 31.4 |
| DeepSort | 453 | 13.02 | 12.17 | 12.58 | 36.13 |
| StrongSort | 397 | 11.85 | 10.87 | 11.34 | 42.53 |
| Sort | 419 | 12.93 | 10.77 | 11.75 | 55.77 |
| DeepMOT | 11 | 74.88 | 81.77 | 78.17 | 60.57 |
| UAVMOT | 16 | 62.48 | 67.67 | 64.97 | 62.43 |
| ByteTrack | 4 | 77.76 | 84.37 | 80.93 | 65.23 |
| BotSort | 0 | 91.46 | 99.93 | 95.51 | 90.6 |
| C_BIoU | 0 | 93.86 | 99.93 | 96.8 | 93.40 |
| **FishMOT** | **0** | **99.97** | **99.97** | **99.97** | **99.93** |

When the number of individuals in a fish school is large, the fish individuals are close to each other. In this stuation, crossing and occlusion phenomena may often occur, which brings great challenges to fish tracking. To verify the performance of these algorithms in this complex scenario, we test them on a 100-fish video, and the results are shown in **Table 2**. The results show that the performance of those classical algorithms on all metrics drops sharply in this complex scenario. DeepSort, DeepMOT, UAVMOT, ByteTrack even have negative values on MOTA, which means that the cumulative number of errors such as FN, FP, and ID_Sw has exceeded the total number of individuals in GT. Even C_BIoU, which performed well on the 10-fish video, has a MOTA of only 23.98%. In contrast, the performance of the method proposed in this paper, FishMOT, does not drop sharply in this complex scenario, and maintains similar performance to that on the 10-fish video on all indicators, with a MOTA of up to 99.06%, which is more than 75% higher than the second best

C_BIoU algorithm. This shows that FishMOT has the ability to handle fish tracking in complex scenarios with a large number of individuals.

Table 2 The comparison with other state-of-the-art MOT algorithms on the 100-fish video

| Methods | IDSw↓ | IDP(%)↓ | IDR(%)↓ | IDF1(%)↑ | MOTA(%)↑ |
|---|---|---|---|---|---|
| CMFTNet | 1691 | 24.21 | 22.3 | 23.22 | 2.34 |
| DeepSort | 2191 | 9.66 | 8.28 | 8.91 | -13.01 |
| StrongSort | 1747 | 21.38 | 18.32 | 19.74 | 0.57 |
| Sort | 1910 | 11.52 | 9.19 | 10.22 | 18.26 |
| DeepMOT | 33 | 46.37 | 47.31 | 46.84 | -3.8 |
| UAVMOT | 206 | 34.32 | 34.49 | 34.41 | -3.14 |
| ByteTrack | 56 | 45.39 | 46.23 | 45.81 | -1.89 |
| BotSort | 21 | 57.29 | 58.57 | 57.93 | 17.04 |
| C_BIoU | 18 | 61.25 | 62.47 | 61.86 | 23.98 |
| **FishMOT** | **0** | **96.52** | **96.54** | **96.54** | **99.06** |

### 4.3 Performance comparison of specialized fish tracking tools

Given that idtracker.ai and Trex are the only currently available tools with visual identification for large groups of individuals, and also because of the quality of results, we will use them as a benchmark for our proposed FishMOT. Results will be compared in terms of Memory, computation time and accuracy, showing FishMOT' ability to achieve a higher level of accuracy typically at far higher speeds, and with a much reduced memory requirement. Here, FishMOT's accuracy is expressed by MOTA. The comparison results are shown in **Table 3**.

Table 3 The comparison with specialized fish tracking tools.

| Methods | Fish_num | Memory↓ | Time(s)↓ | Accuracy↑ |
|---|---|---|---|---|
| idtracker.ai | 10 | 8.26GB | 643s | 99.09% |
|  | 100 | 15.54GB | 1320s | 97.43% |
| TRex | 10 | 4.97GB | 406s | 98.84% |
|  | 100 | 9.37GB | 760s | 97.67% |
| FishMOT(ours) | 10 | **3.76GB** | **72s** | **99.93%** |
|  | 100 | **4.51GB** | **114s** | **99.06%** |

From Table 3, we can see that all these algorithms have high accuracy on both 10-fish and 100-fish videos, FishMOT performs slightly better than the other two on accuracy. However, idtracker.ai and TRex need much more memory space and computation time to complete fish tracking. Especially, the memory space and computation time idtracker.ai need are about 3 times and 9 times those of FishMOT respectively. This is because idtracker.ai uses neural network to extract features for each fish, which requires much computation time and memory usage. TRex uses a similar visual recognition method like idtracker.ai, by adaptively selecting new samples and using different stopping criteria to shorten the training time. So it achieves faster processing speed on the basis of consistent results. Instead, FishMOT is a detection-based multi-object tracking method that builds on YOLOv7 and requires lower processing time and less memory usage. This is because YOLOv7 models the object detection problem as a regression problem, and uses a fully convolutional neural network to predict bounding boxes and class probabilities directly from the input image. This makes its detection time independent of the number of objects in the image, and avoids the need for complex feature extraction or identity assignment per object. So, the performance of FishMOT does not degrade with the increase of the number of fish in the image, unlike other methods that rely on segmentation or classification. In a word, our proposed FishMOT outperforms specialized fish tracking tools, idtracker.ai and TRex.

To test the performance of these algorithms in the situation with uneven illumination, we conduct an experiment by adding uneven illumination elements on 10-fish video. The performance of idtracker.ai and FishMOT under different illumination conditions, shown as **Fig. 7**. the results show that FishMOT is not affected by uneven illumination, while idtracker.ai often fail to track the fish within illumination area. Because idtracker.ai segment fish using background subtraction and it cannot find a suitable threshold to apply. Instead, the proposed FishMOT is a detection-based fish multi-object tracking, which first determines whether the detected fish will perform subsequent tracking. So it will not be affected by other factors, and still maintains the same tracking effect as that of the clear background. It achieves 99.92% MOTA in uneven illumination video. In a word, FishMOT can perform well fish tracking even under complex environment, such as uneven illumination, while idtracker.ai and TRex can not. Idtracker.ai and TRex achieve good results only in the environment with clear background.

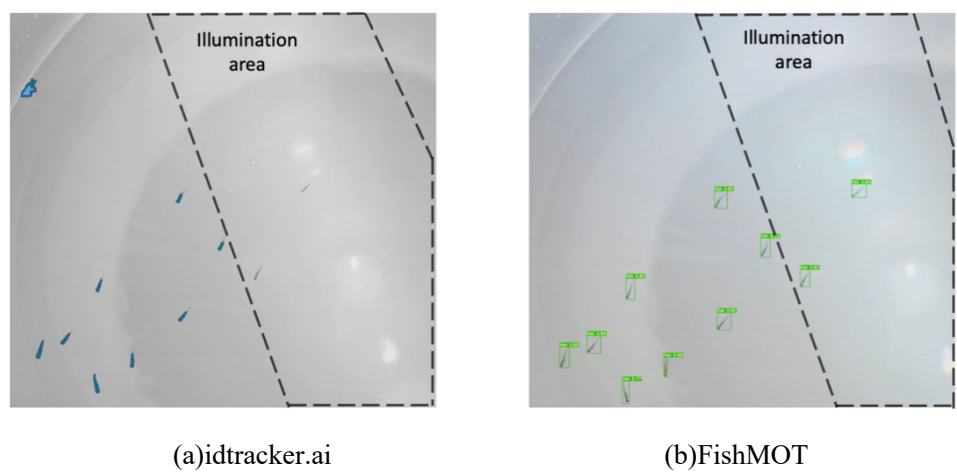

(a)idtracker.ai  (b)FishMOT

**Fig. 7.** The performance of idtracker.ai(Grayscale processing) and FishMOT under uneven illumination conditions.

## 5. Conclusion

In this paper, we proposed FishMOT, a novel fish tracking approach that combines object detection and IoU matching. FishMOT can handle the challenges of fish tracking, such as morphological change, occlusion, and complex environment, by using simple yet effective strategies. FishMOT combines IoU of detection boxes in successive frames and IoU of fish entities to handle the occlusion and morphological change of fish. FishMOT also uses spatio-temporal information to overcome the tracking failure resulting from the missed detection by the detector under complex environment. FishMOT reduces the computational complexity and memory consumption since it does not require complex feature extraction or identity assignment per fish, and does not need Kalman filter to predict the detection boxes of successive frame. We evaluated FishMOT on open-source video datasets and compared it with state-of-the-art multi-object tracking methods and specialized fish tracking tools. The results showed that FishMOT outperformed the existing methods in terms of accuracy, speed, robustness, and generalizability. FishMOT can be applied to various scenarios of fish tracking and behavior analysis in aquaculture and marine ecology.

In future work, we plan to extend FishMOT to handle more complex scenarios of fish tracking, such as underwater or three-dimensional environments. We also plan to explore more applications of FishMOT for fish behavior analysis and aquaculture management.